%% file: template.tex
\documentclass[a4paper]{article}

\usepackage{INTERSPEECH2021}
\usepackage[linesnumbered,algoruled,boxed,lined]{algorithm2e}

\title{Encoder-Decoder Neural Architecture Optimization for Keyword Spotting}
\name{Tong Mo$^1$, Bang Liu$^2$}
\address{
  $^1$University of Alberta, 116 St. and 85 Ave., Edmonton, AB, T6G 2R3, Canada\\
  $^2$University of Montreal, 2900 Edouard Montpetit Blvd, Montreal, Quebec, H3T 1J4, Canada
}
\email{tmo1@ualberta.ca, bang.liu@umontreal.ca}

\begin{document}

\maketitle
\begin{abstract}
Keyword spotting aims to identify specific keyword audio utterances.
In recent years, deep convolutional neural networks have been widely utilized in keyword spotting systems. However, their model architectures are mainly based on off-the-shelf backbones such as VGG-Net or ResNet, instead of specially designed for the task. 
In this paper, we utilize neural architecture search to design convolutional neural network models that can boost the performance of keyword spotting while maintaining an acceptable memory footprint. 
Specifically, we search the model operators and their connections in a specific search space with Encoder-Decoder neural architecture optimization.
Extensive evaluations on Google’s Speech Commands Dataset show that the model architecture searched by our approach achieves a state-of-the-art accuracy of over 97\%.

\end{abstract}

\section{Introduction}

Keyword spotting (KWS) aims to identify certain specific keyword  utterances from speech, such as commands ``yes'', ``no'', ``on'', ``off'', "stop" and "go". Such functionality is critical to many applications, including smart home devices and intelligent cars.  Deploying keyword spotting models on real-world resource-constrained smart devices is a challenging task, as such devices require high keyword spotting accuracy with relatively small number of parameters.

Rapid progress in convolutional neural networks (CNN) triggered research in end-to-end CNN-based models for KWS. Sainath et al. \cite{sainath2015convolutional} introduced CNNs into KWS and showed that CNNs performed well on small footprint keyword spotting.
Since then, multiple off-the-shelf CNN backbones have been widely applied to KWS tasks, such as deep residual network (ResNet) \cite{tang2018deep}, separable CNN \cite{zhang2017hello, majumdar2020matchboxnet, sorensen2020depthwise, xu2020depthwise}, temporal CNN \cite{choi2019temporal} and SincNet \cite{mittermaier2019small}. There are also other efforts to boost performance of CNN models for KWS by combining other deep learning models, such as recurrent neural network (RNN) \cite{arik2017convolutional}, bidirectional long short-term memory (BiLSTM) \cite{zeng2019effective} and streaming layers \cite{rybakov2020streaming}.  However, although the off-the-shelf CNN backbones that existing KWS studies usually relied on have been demonstrated to be effective in image classification tasks, they are not specifically designed for KWS tasks and might not be the perfect architecture for KWS tasks.


Neural network architecture search (NAS) has been widely used to design neural network architectures for specific tasks, such as image classification and language modeling tasks. NAS aims to automate the neural architecture design. Based on the datasets of specific tasks, the model architectures searched by NAS usually achieves better performance for the corresponding tasks compared with human design. Zoph et al. \cite{zoph2016neural} first applied a reinforcement learning approach to search for a neural network architecture for CIFAR-10. Since then, many improved NAS models \cite{xie2018snas, liu2018progressive, zoph2018learning} have appeared, reducing the search space by searching for the best cell and stacking copies of this cell to form a neural network. 
To make the search space continuous, several works \cite{liu2018darts, luo2018neural, chu2020fair} proposed differentiable architecture search approaches, leading to more efficient search through gradient descent or ascent.


Recently, researchers utilized NAS for KWS tasks \cite{Mo2020, zhang2020autokws}. In these research works,  the model architectures were fully designed by DARTS \cite{liu2018darts}, an NAS algorithm which simply assumes the best decision (among different choices of architectures) is the $argmax$ of mixture weights. However, such assumption leads to some biases in the evaluation of the best neural architecture, thus gives sub-optimal performance.

To overcome the aforementioned limitations, in this paper, we leverage Neural Architecture Optimization (NAO) \cite{luo2018neural}, a gradient-based differentiable NAS optimization approach to search for the best convolutional network architecture for KWS.  Specifically, NAO utilizes a decoder to exactly recover the discrete model architecture, which leads to more accurate decision among different architectures. NAO consists of three main components: i) An encoder that maps neural architectures into a continuous space; ii) A predictor that predicts the accuracy of the neural architecture based on the continuous expression from the encoder; and iii) A decoder that maps the continuous expression back to a neural architecture. 

We evaluate the proposed NAO method on the public Google Speech Commands Dataset (Version1 and Version2) \cite{warden2018speech}. 
Our experimental results have shown that the proposed method can find architectures that achieve a state-of-the-art accuracy of over 97\% on both Version1 and Version2 commands dataset, with the setting of 12-class utterance classification, which is the same evaluation setting adopted by most KWS literature \cite{tang2018deep, choi2019temporal,mittermaier2019small, veniat2019stochastic, huh2021metric}.

The rest of this paper is organized as follows. In Section 2 we describe our methods. Experiment results are reported in Section 3. Finally, we conclude in Section 4.

\begin{figure}[!t]
    \centering
    \includegraphics[width = \columnwidth]{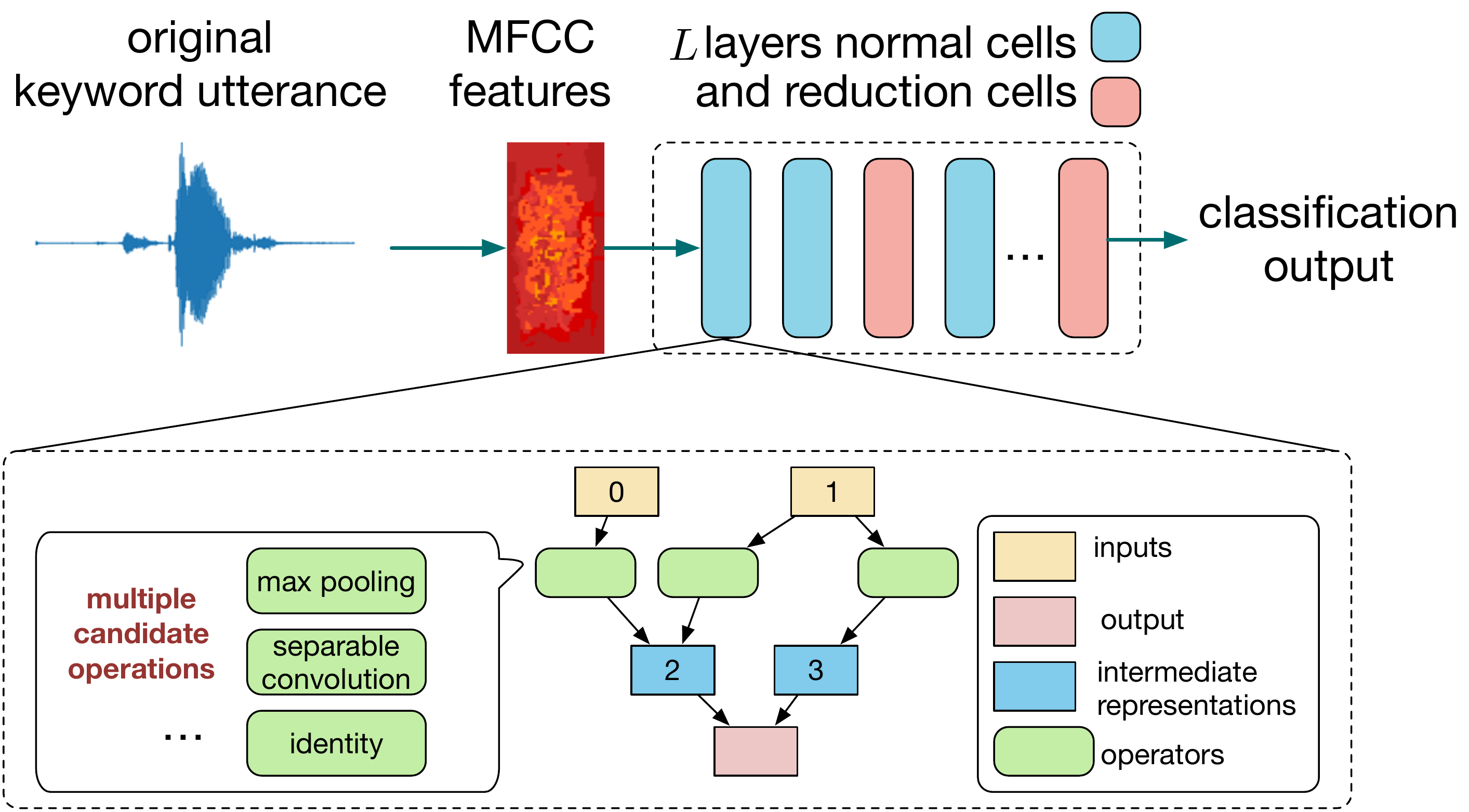}
    \vspace{-2mm}
    \caption{The composition of the convolutional neural network for KWS to be searched by NAO. }
    \vspace{-2mm}
    \label{fig:framework}
\end{figure}

\section{Methods}

As is shown in Figure~\ref{fig:framework}, the CNN we search for is composed of a sequence of $L$ stacked cells, followed by a stem that performs the classification.  The preprocessing procedure to process commands utterances into MFCC features will be described in Section~\ref{sec:experiment}.

We follow the settings of related works \cite{liu2018darts, luo2018neural, Mo2020}, to search for two types of cells: \textit{normal cells} and \textit{reduction cells}. In each cell, there are two input nodes, an output node and several intermediate nodes. The two input nodes of the current cell ($cell_k$) are set equal to the outputs of its previous two cells ($cell_{k-1}$ and $cell_{k-2}$). Each of the intermediate nodes is a latent representation and contains two inputs, which take the outputs of two former nodes, with applying an operation in our search space to it. In order to preserve as many neural architectures as possible, we allow the two inputs of an intermediate node come from the output of one former node, with applying same or different operations to them. The final output node is the concatenation of the outputs of all the nodes that are not consumed by any other nodes. 

\begin{figure*}[!t]
    \centering
    \includegraphics[width = 2\columnwidth]{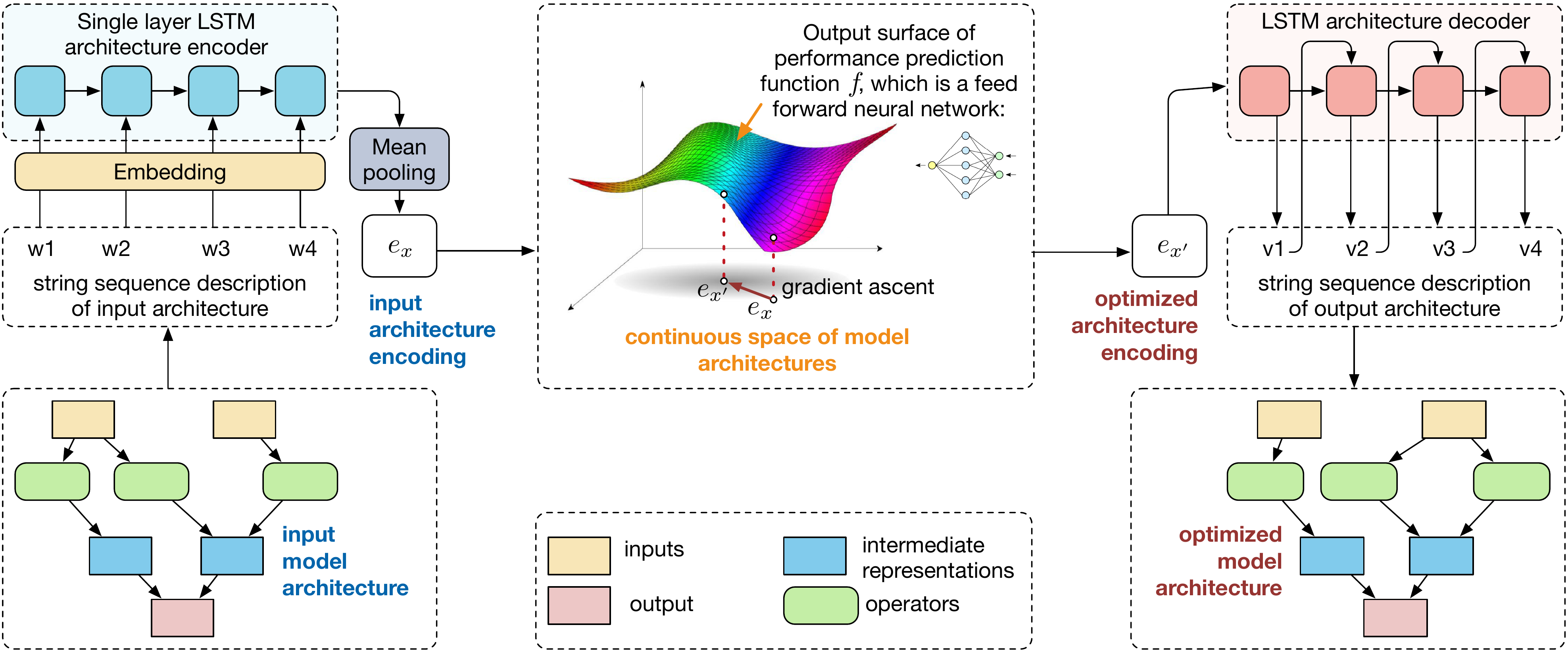}
    \vspace{-2mm}
    \caption{Illustration of Neural Architecture Optimization (NAO)}
     \vspace{-3mm}
    \label{fig:nao}
\end{figure*}

We leverage a differentiable architecture optimization algorithm, NAO \cite{luo2018neural} to explore the search space to find the best cell architectures for KWS. The overall framework of NAO is shown in Figure~\ref{fig:nao}. There are three major parts that NAO consists of: an encoder, a performance predictor and a decoder.

\textit{Encoder}. The function of the Encoder $E$ is to convert a neural architecture $a$ of a cell into a string sequence $x$, and then convert the string sequence into a continuous representation $e_x = E(x)$. Taking the cell in Figure 1 as an example, there are two input nodes, node 0 and node 1, and two intermediate nodes, node 2 and node 3. The input edge of each intermediate node is represented via three tokens, including the node index it selected as the input, the operation type and operation size. For node 2,  a string "node 0 sep-conv 3x3 node 1 max-pooling 3x3"  represents node 2 takes the output of node 0 and node 1 as inputs, and respectively apply 3 × 3 separable convolution and 3 × 3 max pooling. We use a single layer Long short-term memory (LSTM) as the encoder, which takes the string sequence as inputs and outputs the hidden states of the LSTM. The hidden states are continuous representations of the architecture.

\textit{Performance predictor}. The Performance predictor $f$ maps the continuous representation $e_x$ into its performance $s_x$ based on the evaluation on the validation dataset. The predictor is constituted by a mean pooling layer and a feed forward network. To train this predictor, the predictor least square regression loss $\mathcal{L}_{pred}$ for the all candidate architectures set $\mathcal{X}$ is minimized:
\begin{equation}
\label{eqn:predictor}
\begin{aligned}
\min&\quad \mathcal{L}_{pred}=\sum_{x \in \mathcal{X}}(s_x - f(E(x)))^2. \\
\end{aligned}
\end{equation}

\textit{Decoder}. The decoder $D$ maps the continuous representation $e_x$ back to the string sequences $x$ corresponding to an architecture $a$. We set $D$ as an LSTM model with the initial hidden state. An attention mechanism \cite{bahdanau2014neural} is leveraged to optimize the decoder, which will output a context vector at each time step $r$. The decoder D induces a factorized distribution which is identical to the one in \cite{luo2018neural}:

\begin{equation}
\label{eqn:decoder}
\begin{aligned}
P_D(x|e_x) = \prod_{r=1}^{T} P_D(x_r|e_x, x_{<r}),\\
\end{aligned}
\end{equation}
where $T$ is the total number of time steps of the encoder. To train the decoder, the structure reconstruction loss $\mathcal{L}_{rec}$ is minimized:

\begin{equation}
\label{eqn:decoderloss}
\begin{aligned}
\min&\quad \mathcal{L}_{rec}=-\sum_{x \in \mathcal{X}} \log P_D(x|E(x)). \\
\end{aligned}
\end{equation}

The encoder, performance predictor and decoder are jointly trained by minimizing the weighted loss sum $\mathcal{L}$:

\begin{equation}
\label{eqn:totalloss}
\begin{aligned}
\min&\quad \mathcal{L}=\lambda \mathcal{L}_{pred} + (1-\lambda) \mathcal{L}_{rec} , \\
\end{aligned}
\end{equation}
where $\lambda$ is a float hyper parameter between 0 and 1. The detailed algorithm is shown in Alg~\ref{algo:nao}.
\begin{algorithm}[tp] \small 
    \caption{Encoder-Decoder Neural Architecture Optimization for Keyword Spotting}\label{algo:nao}
    \textbf{Input}: Initial candidate architectures set $\mathcal{X}$ with legal $x$, evaluation architecture set $\mathcal{X}_{eval}=\mathcal{X}$,  performances set $S = \emptyset$, number of iterations $N$. \\
    \For{$n = 1,..., N$} {
    	Train each architecture $x \in \mathcal{X}_{eval}$, and evaluate architectures on validation set to get a validation performances set $S_{eval} = \{s_x\}$\;
    	Enlarge $S$: $S = S \cup S_{eval}$\;
    	Train $E$, $f$ and $D$ by minimizing Eqn~\ref{eqn:totalloss}\;
    	Obtain better continuous representations $e_x$ based on the $E$ and $f$. Form a representations set $EX = \{e_x\}$\;
    	Decode all continuous representations in $EX$ back to architectures. Update $\mathcal{X}_{eval} = \{D(e_x), \forall e_x \in EX\}$. Enlarge $\mathcal{X} = \mathcal{X} \cup \mathcal{X}_{eval}$\;
    }
    \Return the best-performance architecture in $\mathcal{X}$\;
\end{algorithm}

After the search ends, the normal and reduction cell architectures with the best validation performance are returned respectively. To form the final CNN architecture by stacking the cells, we locate the reduction cells at the 1/3 and 2/3 of the total depth of the network.

\section{Performance Evaluation}
\label{sec:experiment}

We evaluate the proposed method for keyword spotting on two Google Speech Commands Dataset versions, Version1 and Version2 \cite{warden2018speech}. Version1 contains 65,000 one-second-long audio utterances pertaining to 30 words, while Version2 contains 105,000 one-second-long audio utterances pertaining to 35 words. There are approximately 2,200 and 3000 samples for each word in Version1 and Version2 respectively. Following the same setting as \cite{tang2018deep, choi2019temporal}, we cast the problem as a classification task that distinguishes among 12 classes, i.e., ``yes'', ``no'', ``up'', ``down'', ``left'', ``right'', ``on'', ``off'', ``go'', ``stop'', an unknown class, and a silence class. These ten words are acknowledged useful commands in the Internet of things (IoT) and robotics applications \cite{warden2018speech}. The unknown class contains utterances sampled from the remaining words other than the above ten words, while the silence class has utterances with only background noise. The entire dataset is split into 80\% training, 10\% validation, and 10\% testing sets. The training set and validation set are used during architecture search optimization, and are further combined to form a new training set for evaluating the best architecture on the test set.

\subsection{Audio preprocessing}
We follow the identical audio utterances preprocessing procedure of multiple small-footprint KWS works \cite{sainath2015convolutional,tang2018deep,choi2019temporal,Mo2020, huh2021metric}. We construct forty-dimensional Mel-Frequency Cepstrum Coefficient (MFCC) frames for each audio utterance and stack them using a 30-millisecond window size and a 10-millisecond time frame shift.

\subsection{Search settings}
We set the token embedding size and hidden state size of the LSTM in the encoder to 32 and 96 respectively. The hidden states of LSTM are normalized to unit length. The hidden state size of the LSTM in the decode is set as 96.  We train the encoder, performance predictor and decoder of NAO using Adam optimization algorithm for 1000 epochs with $\lambda=0.9$ and a learning rate of 0.001. During neural architecture optimization, we set both the number of iterations $N$ and the number of cells to 3 and train each found architecture during the search for 25 epochs. The batch size is set to 128 and the number of channels is set to 16. We search cells for Version1 and Version2 dataset respectively with the identical hyper-parameters settings.

Inside each cell, we set the number of intermediate nodes to 5. In order to evaluate the performance of NAO fairly, our search space does not contain high-performance CNN blocks from other KWS works. The search space only consists of five simple candidate operations: identity, $3 \times 3$ separable convolution, $5 \times 5$ separable convolution, $3 \times 3$ average pooling, $3 \times 3$ max pooling.
All separable convolutions are always applied twice and follow ReLU-Conv-BN (Batch Normalization) order \cite{zoph2018learning, liu2018darts}.

\subsection{Evaluation settings}

During the evaluation, we instantiate the network to be tested based on the best cell architecture with the highest validation accuracy found by the search phase for Version1 and Version2 respectively, and experiment with the CNN formed by stacking the cells. We name our models NAO1 and NAO2, which correspond to the Version1 and Version2 datasets respectively. We randomly re-initialize the weights in the network and re-train it from scratch for 200 epochs to report the evaluation results. We evaluate the models discovered for Version1 and Version2 in terms of the accuracy and model size, with 12 and 16 initial channels respectively. We compare to the following baseline models which are mostly KWS models based on CNN \footnote{MatchboxNet-$3\times2\times64$ \cite{majumdar2020matchboxnet} achieves state-of-the-art results on Google Speech Dataset Version1. It trains its model on a self-rebalanced datset, so the results are not comparable to ours. We also include the state-of-the-art LSTM models \cite{zeng2019effective}.
}:




\begin{itemize}
    \item \textbf{Res15}: the best ResNet variant based on regular convolutions in \cite{tang2018deep}.
    \item \textbf{Attention RNN}: the best model reported in \cite{de2018neural}, which is a convolutional recurrent network model with attention for KWS.
    \item \textbf{TC-ResNet14-1.5}: a ResNet variant achieving the highest accuracy in \cite{choi2019temporal}.
    \item \textbf{SincConv+DSConv}: the best model reported in \cite{mittermaier2019small}, which first uses the Sinc-convolution to extract features from raw audio.
    \item \textbf{DenseNet-BiLSTM}: the best model reported in \cite{zeng2019effective} following data preprocessing procedures in \cite{tang2018deep, choi2019temporal, mittermaier2019small}. 
    \item \textbf{NAS2}: the best model reported in \cite{Mo2020}, which first applies DARTS \cite{liu2018darts} to search CNN models for KWS.
    \item \textbf{NoisyDARTS-TC14}: the best model reported in \cite{zhang2020autokws}, which applies NoisyDARTS \cite{chu2020noisy} to search CNN models for KWS based on blocks designed in \cite{choi2019temporal}.

\end{itemize}


\subsection{Evaluation results}
\begin{figure}[t]
    \centering
    \includegraphics[width = \columnwidth]{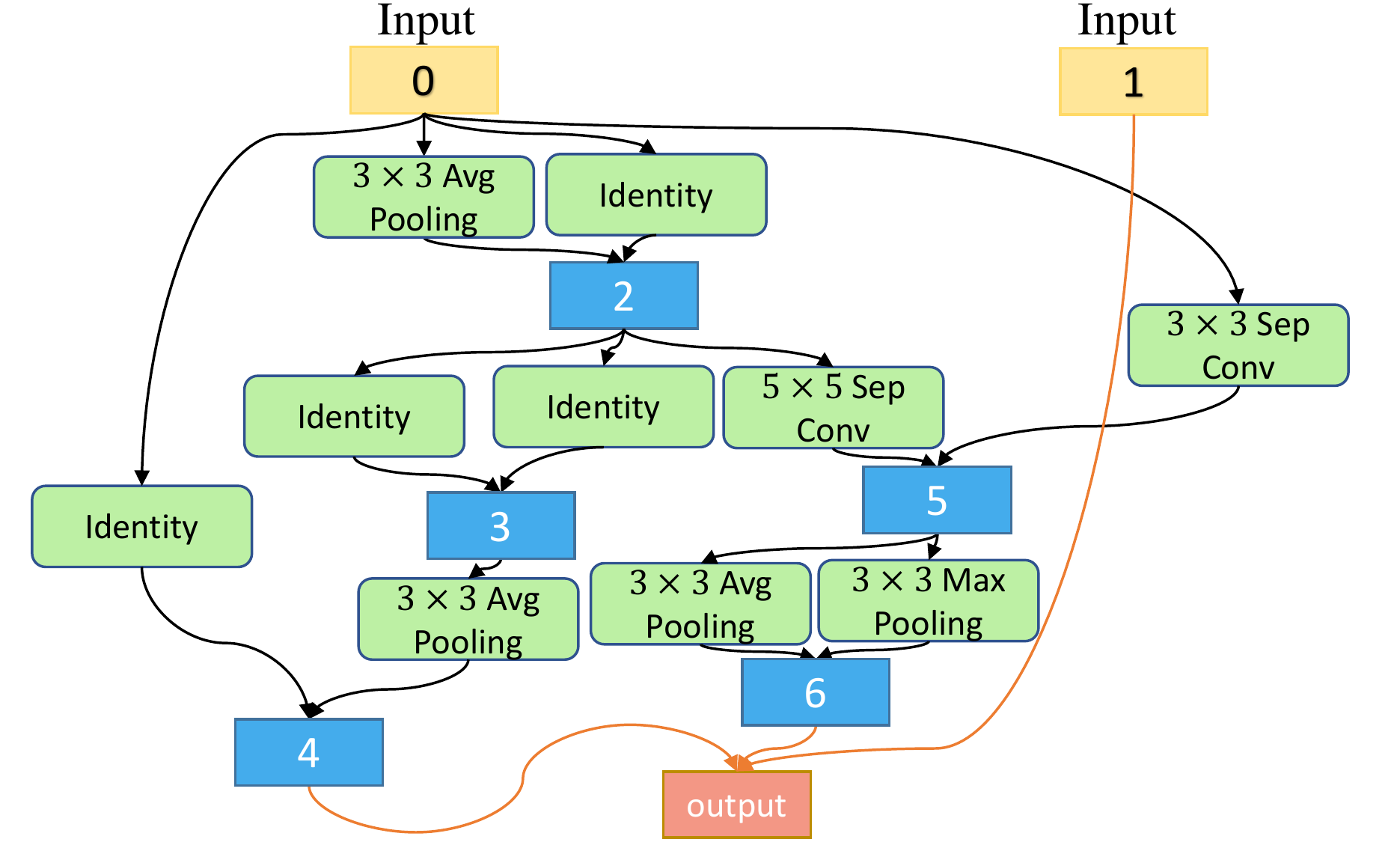}
    \vspace{-5.5mm}
    \caption{The normal cell found for Google Commands Dataset Version1.}
    \vspace{-1mm}
    \label{fig:normal1}
\end{figure}

\begin{figure}[t]
    \centering
    \includegraphics[width = \columnwidth]{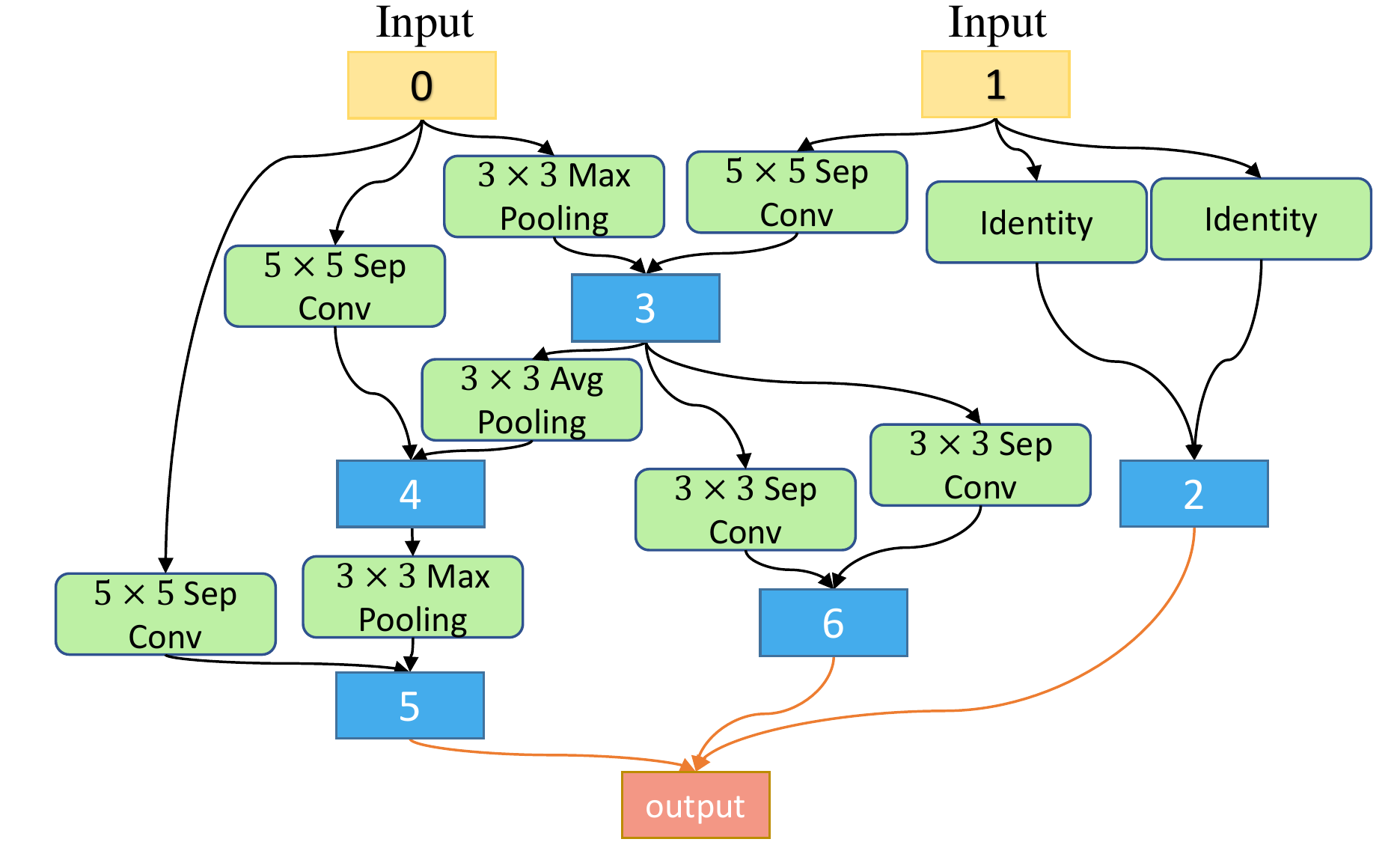}
    \vspace{-5.5mm}
    \caption{The reduction cell found for Google Commands Dataset Version1.}
    \vspace{-1mm}
    \label{fig:reduction1}
\end{figure}

\begin{figure}[t]
    \centering
    \includegraphics[width = \columnwidth]{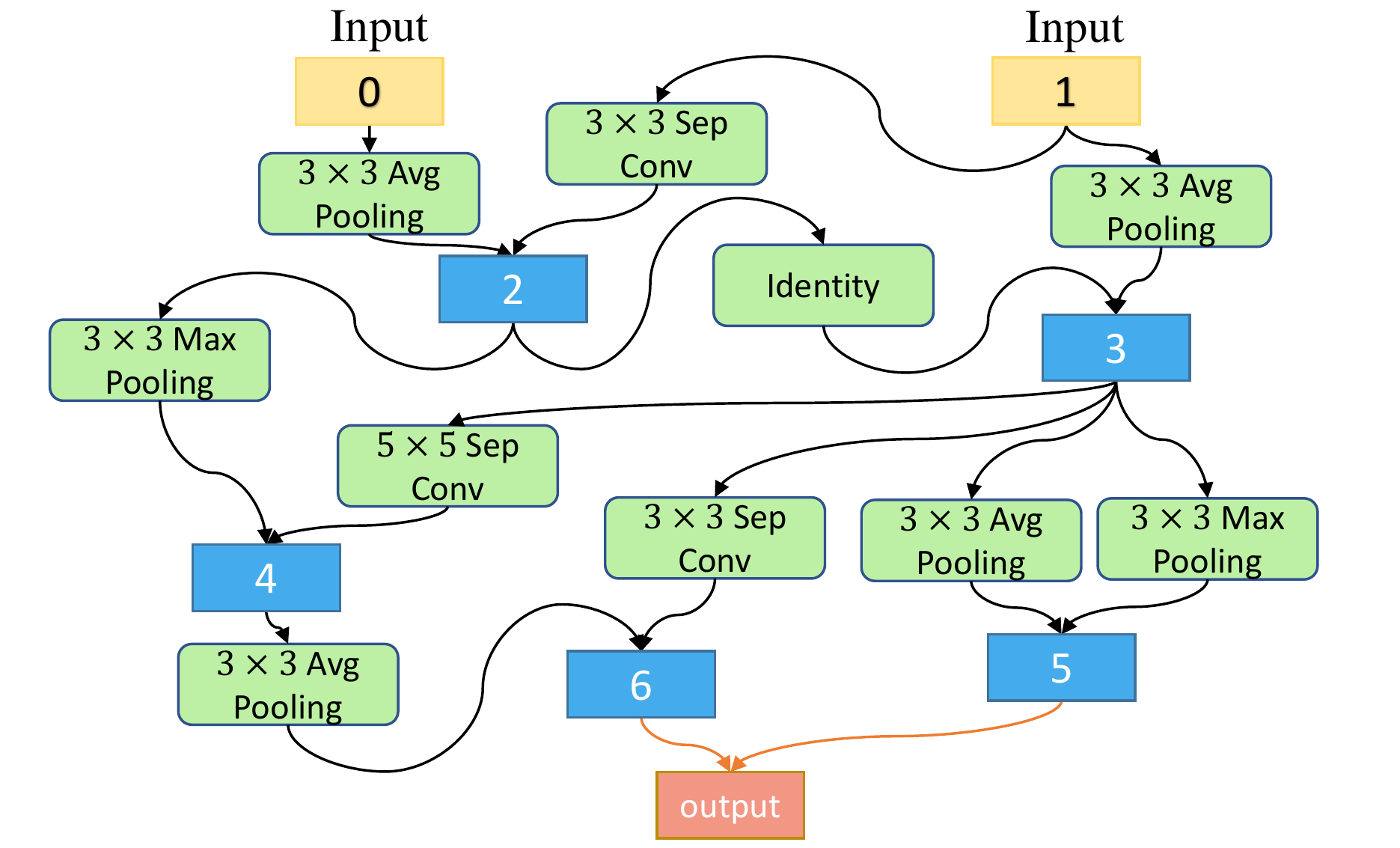}
    \vspace{-5.5mm}
    \caption{The normal cell found for Google Commands Dataset Version2.}
    \vspace{-1mm}
    \label{fig:normal2}
\end{figure}

\begin{figure}[t]
    \centering
    \includegraphics[width = \columnwidth]{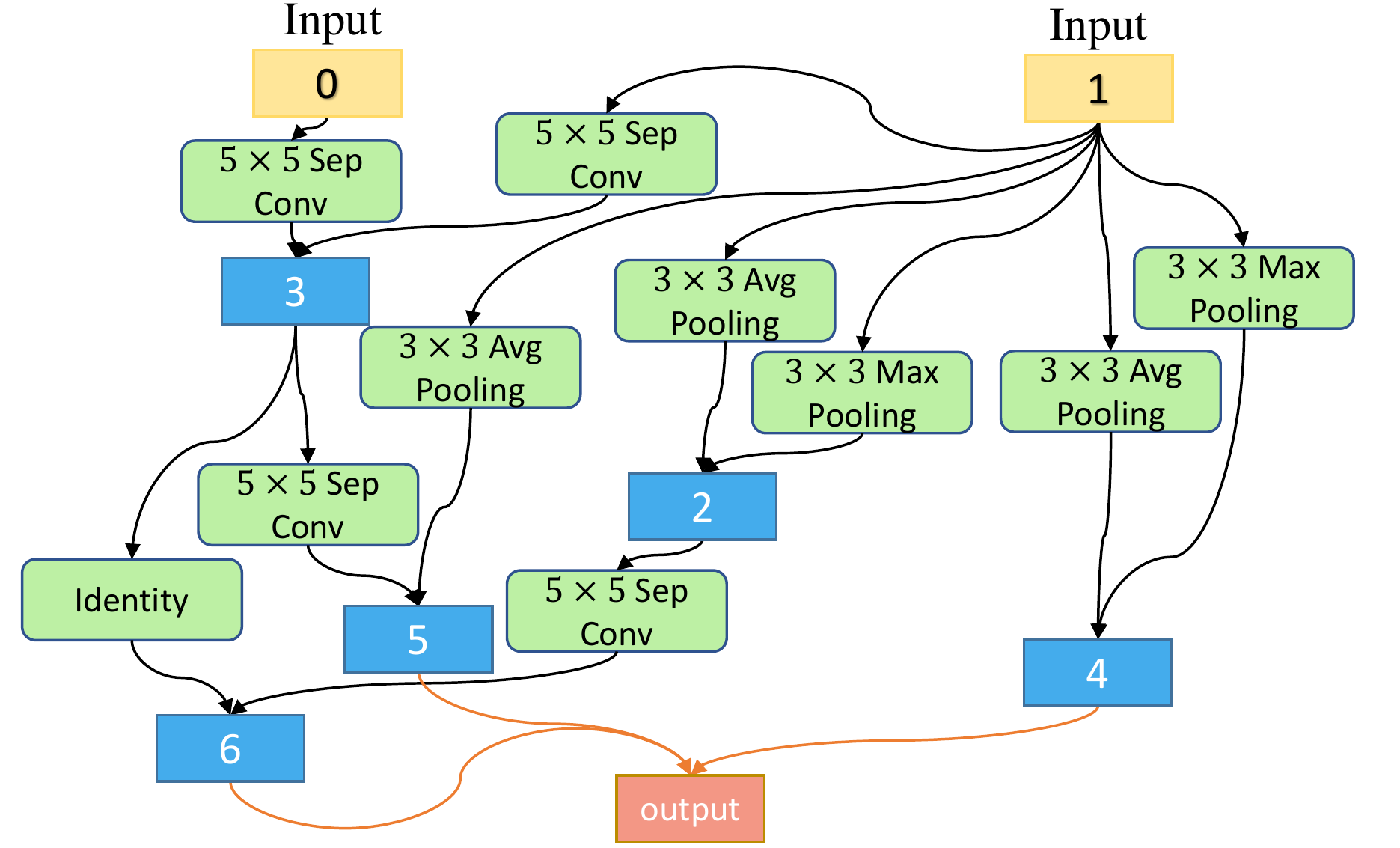}
    \vspace{-5.5mm}
    \caption{The reduction cell found for Google Commands Dataset Version2.}
    \vspace{-1mm}
    \label{fig:reduction2}
\end{figure}

Figures~\ref{fig:normal1}-\ref{fig:reduction2} illustrate the cells found for Google Speech Commands Dataset Version1 and Version2. The search costs are 8 GPU days and 10 GPU days respectively on an NVIDIA GPU (Tesla V100). Table 1 shows a performance comparison between our models and baseline models on Version1 dataset. From this table, we can observe that the NAO1 model with 12 initial channels outperforms Res15, Attention RNN, TC-ResNet, SincConv and DenseNet-BiLSTM in terms of accuracy with a similar number of parameters. And NAO1 with 16 channels can achieve a new state-of-the-art accuracy of 97.28\% with an acceptable model size of 469K parameters.

Results on Version2 dataset in Table 2 again show that our approach can find competitive CNN compared with other baselines. Two NAO2 models, with 12 and 16 initial channels respectively, can outperform all baseline models with acceptable parameter sizes. Notably, NAO2 with 16 channels achieves a new state-of-the-art accuracy of 97.92\% on Version2 dataset.



\input{Table}
Moreover, NAS2 \cite{Mo2020} and NoisyDARTS-TC14 \cite{zhang2020autokws} are state-of-the-art KWS models obtained by neural architecture search methods. NAO1 can outperform NAS2 in terms of both accuracy and the number of parameters on Version1 dataset, which shows the strong advantages of our proposed methods.
In addition, NAO1 and NAO2 can beat NoisyDARTS-TC14 on the Version1 and Version2 datasets respectively. Although the model sizes are not as small as NoisyDARTS-TC14, the main reason is that NAO does not depend on high-performance small footprint blocks designed in \cite{choi2019temporal}, and only use common operations to constitute the search space. Based on the comparison with NAS2 \cite{Mo2020} and NoisyDARTS-TC14 \cite{zhang2020autokws}, it confirms that NAO variants have improved searching capability with the encoder-decoder optimization framework.

\input{Table2}
\vspace{-5.5pt}
\section{Conclusions}

While most existing keyword spotting models rely on manually designed convolutional neural networks, neural architecture search has been applied to find better CNN models for keyword spotting. In this paper, we propose an improved neural architecture search method for keyword spotting, named as encoder-decoder neural architecture optimization to search for CNN architectures for keyword spotting.
The encoder, performance predictor and decoder jointly make the search algorithm more effective to discover better CNN architectures for keyword spotting. The architectures discovered by our proposed approach achieve state-of-the-art results on the Google Speech Command Dataset Version1 and Version2.

\bibliographystyle{IEEEtran}

\bibliography{mybib}


\end{document}

%% file: Table.tex
\begin{table}[!t]
\centering
\caption{ Comparison with state-of-the-art models
on Google Speech Command Dataset Version1. The numbers marked with $\dagger$ are taken from the corresponding papers. '-' means not available. The best result is marked in bold.}

\begin{tabular}{lccc}

\hline

\textbf{Model}    
  &
 \textbf{\begin{tabular}[c]{@{}c@{}} Channels \\(\#)\end{tabular}}
 &
 \textbf{\begin{tabular}[c]{@{}c@{}} Acc.\\ (\%)\end{tabular}}    & \textbf{\begin{tabular}[c]{@{}c@{}} Par.\\(K)\end{tabular}}
\\\hline
Res15 \cite{tang2018deep}  & -         & 95.8$\dagger$                                                & 239     \\ 
Attention RNN \cite{de2018neural} & -    & 95.6$\dagger$                                                & 202       \\
TC-ResNet14-1.5 \cite{choi2019temporal} & -    & 96.6$\dagger$                                                & 305       \\
SincConv+DSConv \cite{mittermaier2019small}                 & -        & 96.6$\dagger$                      & 122     \\ 
DenseNet-BiLSTM \cite{zeng2019effective}                 & -        & 96.2$\dagger$                      & 223     \\ 
NAS2 \cite{Mo2020}                & -        & 97.2$\dagger$                      & 886     \\                  NoisyDARTS-TC14 \cite{zhang2020autokws}                & -        & 97.2$\dagger$                      & 109     \\                    
\hline

NAO1   & 
\begin{tabular}[c]{@{}c@{}}  12 \\ 16 \end{tabular}     & 
\begin{tabular}[c]{@{}c@{}}  97.01 \\ \textbf{97.28} \end{tabular}     & 
\begin{tabular}[c]{@{}c@{}}  278 \\ 469 \end{tabular}  \\   
\hline

\end{tabular}

\end{table}

%% file: Table2.tex
\begin{table}[!t]
\centering
\caption{ Comparison with state-of-the-art models
on Google Speech Command Dataset Version2. }

\begin{tabular}{lccc}

\hline

\textbf{Model}    
  &
 \textbf{\begin{tabular}[c]{@{}c@{}} Channels \\(\#)\end{tabular}}
 &
 \textbf{\begin{tabular}[c]{@{}c@{}} Acc.\\ (\%)\end{tabular}}    & \textbf{\begin{tabular}[c]{@{}c@{}} Par.\\(K)\end{tabular}}
\\\hline
Attention RNN \cite{de2018neural} & -    & 96.9$\dagger$                                                & 202       \\
DenseNet-BiLSTM \cite{zeng2019effective}                 & -        & 97.3$\dagger$                      & 223     \\ 
NoisyDARTS-TC14 \cite{zhang2020autokws}                & -        & 97.4$\dagger$                      & 107     \\                    
\hline

NAO2   & 
\begin{tabular}[c]{@{}c@{}}  12 \\ 16 \end{tabular}     & 
\begin{tabular}[c]{@{}c@{}}  97.50 \\ \textbf{97.92} \end{tabular}     & 
\begin{tabular}[c]{@{}c@{}}  274 \\ 458 \end{tabular}  \\   
\hline

\end{tabular}

\end{table}